\documentclass[10pt,twocolumn,letterpaper]{article}

\usepackage[pagenumbers]{cvpr} 

\def\cvprPaperID{**}
\def\confName{****}
\def\confYear{****}

\usepackage{graphicx}
\usepackage{amsmath}
\usepackage{amssymb}
\usepackage{booktabs}
\usepackage{multirow}

\usepackage{fp}
\usepackage{expl3}[2012-07-08]
\ExplSyntaxOn
\ExplSyntaxOff

\usepackage{amsmath,amsfonts,bm}

\def\x{{x}}

\def\xi{{\x_i}}

\newcommand{\ignorethis}[1]{}

\def\eqref#1{equation~\ref{#1}}

\def\1{\bm{1}}

\DeclareMathAlphabet{\mathsfit}{\encodingdefault}{\sfdefault}{m}{sl}
\SetMathAlphabet{\mathsfit}{bold}{\encodingdefault}{\sfdefault}{bx}{n}

\usepackage{graphicx}
\usepackage{amsmath}
\usepackage{xcolor}
\usepackage{soul}
\usepackage{caption}
\usepackage{subcaption}

\usepackage[pagebackref,breaklinks,colorlinks]{hyperref}

\usepackage[capitalize]{cleveref}
\crefname{section}{Sec.}{Secs.}
\Crefname{section}{Section}{Sections}
\Crefname{table}{Table}{Tables}
\crefname{table}{Tab.}{Tabs.}

\begin{document}

\newcommand{\sysName}{StyleRetoucher}

\newcommand{\wanchao}[1]{{\color{blue} #1}}
\newcommand{\wanchaos}[1]{{\color{blue} [Wanchao: #1]}}
\newcommand{\hbc}[1]{{\color{red} [HB: #1.]}}
\newcommand{\hb}[1]{{\color{cyan} #1}}
\newcommand{\lj}[1]{{\color{teal} [LJ: #1.]}}
\newcommand{\wangcan}[1]{{\color{teal} [Wangcan: #1]}}

\title{\sysName: {Generalized} Portrait Image Retouching with GAN Priors}

\author{Wanchao Su \\
Monash University \\
{\tt\small wanchao.su@monash.edu}
\and
Can Wang \\
City University of Hong Kong \\
{\tt\small cwang355-c@my.cityu.edu.hk}
\and
Chen Liu \\
Nanyang Technological University \\
{\tt\small chen034@ntu.edu.sg}
\and
Fangzhou Han \\
City University of Hong Kong \\
{\tt\small han.fangzhou@my.cityu.edu.hk}
\and
Hongbo Fu \\
City University of Hong Kong \\
{\tt\small hongbofu@cityu.edu.hk}
\and
Jing Liao\footnotemark[1] \\
City University of Hong Kong\\
{\tt\small jingliao@cityu.edu.hk}
}

\maketitle

{
  \renewcommand{\thefootnote}%
    {\fnsymbol{footnote}}
  \footnotetext[1]{Corresponding Author.}
}

\begin{abstract}
Creating fine-retouched portrait images is tedious and time-consuming even for professional artists. There exist automatic retouching methods, but they either suffer from over-smoothing artifacts or lack generalization ability. To address such issues, we present \sysName, a novel automatic portrait image retouching framework, leveraging StyleGAN’s generation and generalization ability to improve an input portrait image's skin condition while preserving its facial details. Harnessing the priors of pretrained StyleGAN, our method shows superior robustness: a). performing stably with fewer training samples and b). generalizing well on the out-domain data. Moreover, by blending the spatial features of the input image and intermediate features of the StyleGAN layers, our method preserves the input characteristics to the largest extent. We further propose a novel blemish-aware feature selection mechanism to effectively identify and remove the skin blemishes, improving the image skin condition. Qualitative and quantitative evaluations validate the great generalization capability of our method. Further experiments show ~\sysName's superior performance to the alternative solutions in the image retouching task. We also conduct a user perceptive study to confirm the superior retouching performance of our method over the existing state-of-the-art alternatives.
\end{abstract}

\section{Introduction}\label{sec:intro}
Portrait images are in great need for various usage scenarios, such as fashion, advertising, {figure-branding}, etc. Raw images often need to be retouched through a series of operations, among which the most important procedure is to remove 
skin blemishes (e.g., acnes, scars, uneven skin color/texture, extreme reflection, etc.). Manually retouching a face photo is time-consuming and labor-intensive, {and also requires long-time learning,} hindering people from getting well-retouched portraits effortlessly.

\begin{figure}[t]
    \centering
    \includegraphics[width=0.9\linewidth]{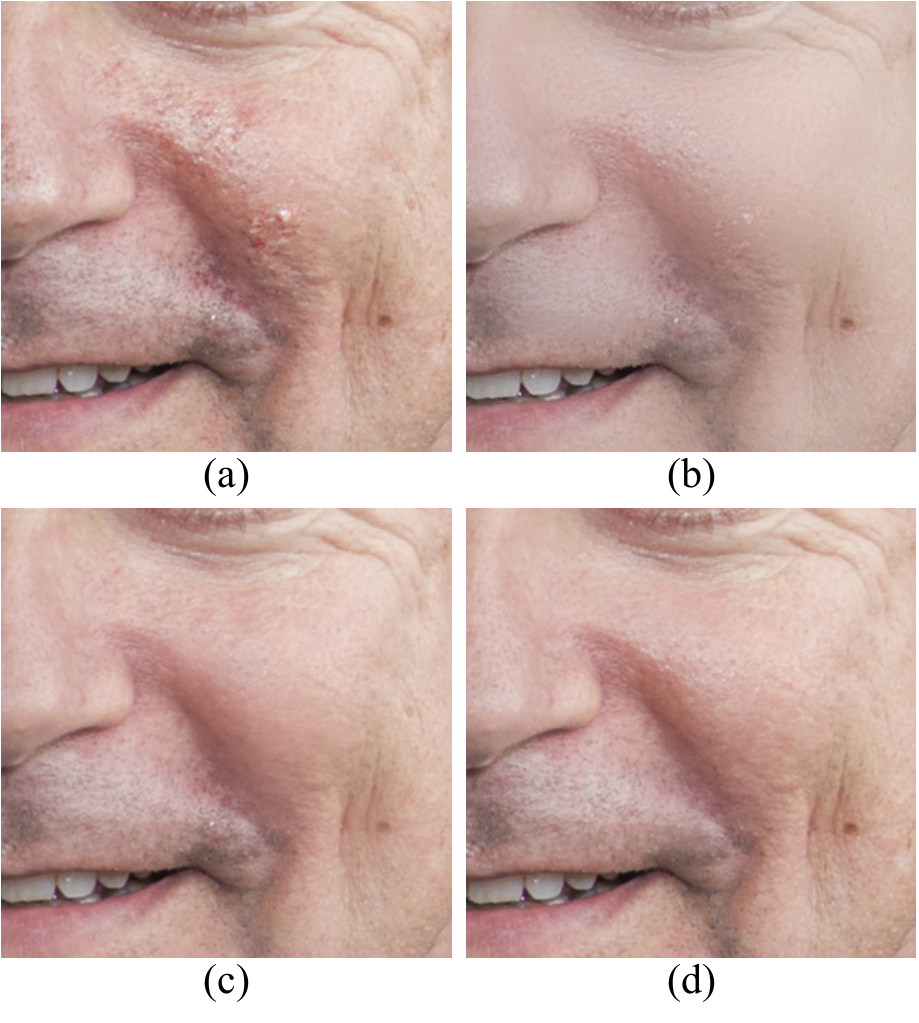}
    \caption{Retouching effect comparison. (a): input, (b): BeautyCam~\cite{beautycam2013meitu}, (c): our method, and (d): a manually retouched result by a professional artist. 
    }
    \label{fig_retouch_effects}
\end{figure}

\begin{figure*}
    \centering
    \includegraphics[width=0.98\linewidth]{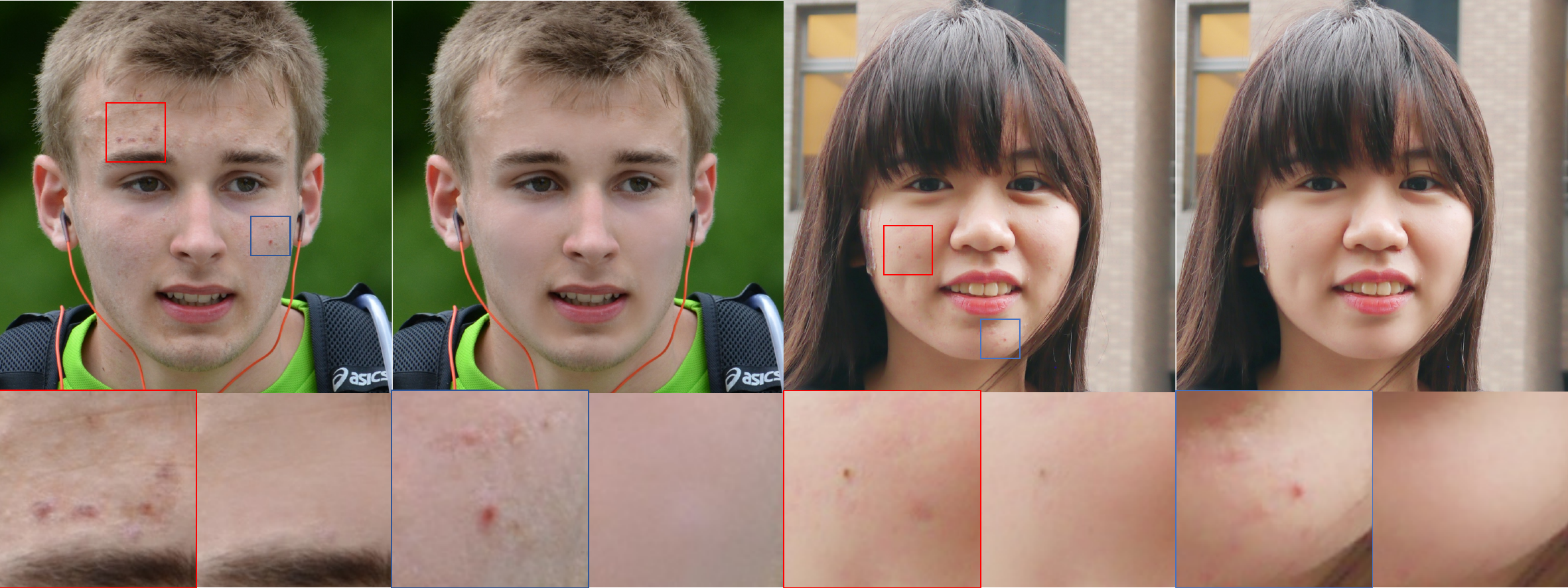}
    \caption{
    Our \sysName~automatically removes skin blemishes in portrait photos and preserves the characteristics (i.e., fine details) of the original inputs (Left), producing high-quality photo retouching results (Right). Four pairs of close-ups in the bottom row show the regions before and after retouching. Please zoom in to examine the results in detail.} 
    \label{fig_teaser}
\end{figure*}

Various approaches have attempted to achieve automatic image retouching, which essentially involves two steps: first, identify and remove blemish areas of an input image and then inpaint compatible content to the blemish areas while maintaining the rest of the image intact. Light-weight retouching procedures, often implemented as mobile applications like BeautyCam~\cite{beautycam2013meitu}, apply various filters to smooth out skin blemishes. However, the results often exhibit over-smoothing effects, erasing the details {(e.g., skin textures, moles, etc.)} of the input{, introducing undesired whitening effect,} and presenting noticeable gaps with the manually retouched results by experienced artists (see Figure \ref{fig_retouch_effects} for an example). Recent efforts toward image retouching adopt learning-based approaches.  They are usually based on deep learning models to achieve retouching effects ~\cite{Shafaei2021WACV,lei2022abpn}.  Such methods suffer from limited generalization capability (i.e., unable to handle data different from that used in training settings, {incorporating undesirable content (e.g., skin textures)} or over-smoothing results due to the erasing of the important input details). {Since the retouching preference highly represents the retoucher's style and well-retouched result to the input is expensive to obtain.  It would be beneficial for both retouching professionals and ordinary people to develop a more versatile retouching method that can learn from smaller datasets and apply that knowledge effectively in various scenarios.}

To address these issues, we propose \sysName, the first GAN-based retouching framework that utilizes pre-trained StyleGAN's superior generation and generalization ability to retouch portrait photos automatically. StyleGAN\cite{karras2019style}-based architectures efficiently solve portrait-centered problems, such as super-resolution \cite{menon2020pulse,chan2021glean}, restoration \cite{wang2021towards,wang2022panini,yang2021gan}, inpainting \cite{abdal2020image2styleganpp,richardson2020encoding}, attribute manipulation \cite{patashnik2021styleclip,harkonen2020ganspace,abdal2021styleflow,gal2022stylegan}, and stylization \cite{chong2022jojogan,song2021agilegan,pinkney2020resolution,cao2018cari}, to name a few. StyleGAN, pre-trained on a large portrait dataset, generates facial content with simple control and high quality.  Thus various methods adopt it as the generation backbone in producing desired imagery content. Likewise, we find StyleGAN a perfect fit for retouching when we infuse the input to the StyleGAN generation process, utilizing its ability in producing desired skins.
{See Section \ref{sec_gpp} for detailed elaboration.} The incorporation of StyleGAN provides superior generalization capability to our framework, enabling the challenging retouching for a). out-domain cases and 2). stable retouching ability with fewer training samples.   By adding a simple yet effective blemish-aware feature selection module, we successfully couple the input semantics with the StyleGAN priors, producing retouching results with detail preservation to the largest extent while removing skin blemishes at the same time. {We show two demonstrative examples of our method in Figure \ref{fig_teaser}.}

\begin{figure}[t]
    \centering
    \includegraphics[width=\linewidth]{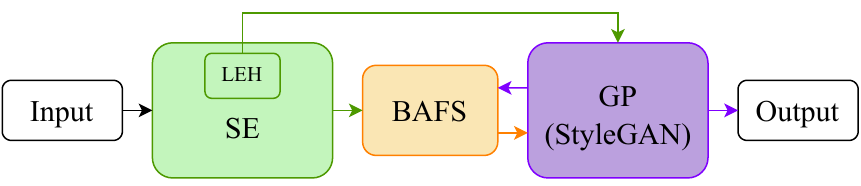}
    \caption{Overview of our \sysName~pipeline. Our method preserves portrait details through cascaded feature blending between StyleGAN (GAN Prior, GP) and the input photo (Semantic Extraction, SE) using Blemish-Aware Feature Selection (BAFS). To extract the high-level semantics, we propose Latent Extraction Head (LEH) on top of the SE module.}
    \label{fig_intro_pipeline}
\end{figure}

{The high-level overview of our \sysName~is illustrated} in Figure \ref{fig_intro_pipeline}: To preserve the portrait photo's details to the largest extent, we adopt a spatially cascaded feature matching between the pre-trained StyleGAN generation process (see GAN Prior (GP) Module in Section \ref{sec_gpp}) and the original input portrait. We propose a Semantic Extraction Module (SE in Section \ref{sec_sfp}) to extract the spatial features of the input photo in a cascaded manner {and thus to} better incorporate input semantics. Meanwhile, we also propose a latent extraction head (LEH) to invert the input photo to the StyleGAN $W^+$ space for controlling the generation of StyleGAN, guided by the self-adaptive weights in the generation process. Our Blemish-Aware Feature Selection Module (BAFS, Section \ref{sec_cafb}) generates the spatial- and channel-wise masks for the pyramids of spatial feature maps produced by the SE and GP modules, which implicitly identify and remove the input skin blemishes in the feature space, enabling a smooth data transition from the input to the output. We constructed a new set of data based on the data pairs from Flickr-Faces-HQ (FFHQ) \cite{karras2019style} and Flickr-Faces-HQ-Retouching (FFHQR) dataset \cite{Shafaei2021WACV} to train our model in an end-to-end way. 

Our \sysName~for the first time leverages {a pre-trained} StyleGAN for the image retouching task and achieves outstanding performance. The integration of the pre-trained StyleGAN enhances the generalization of our framework, since we convert the input-output mapping problem to finding the closest sample in GAN prior space procedure. Due to the design of the BAFS module and cascaded incorporation of input semantics, our method produces retouched results with blemishes removed and fine details preserved. {Although the StyleGAN is trained with a large volume of general face dataset, incorporating such a pre-trained model in our framework significantly alleviates the dependence on large-scale \emph{paired well-retouched} datasets, which are, in contrast, essential for current deep learning-based retouching methods.} We have conducted thorough quantitative and qualitative evaluations to demonstrate its superior performance over alternative solutions. We show our framework's generalization capability in two ways: a) outperforming the alternative solutions on out-domain data by a large margin; b) functioning well even on a small scale of training data. We also performed a perceptive study by inviting a group of participants to rank the results in terms of retouching quality, and the results further confirmed our method's outstanding performance. Our method further supports changing the strength of retouching in a content-aware manner.

\section{Related Works}

Our method is closely related to the topics of skin image retouching and GAN prior incorporation. For each topic, we only discuss the most related works to ours; a comprehensive survey of the concerning topics is beyond the scope of this paper.

\subsection{Skin Image Retouching}
One trend of existing image retouching works aims to adjust the coloring scheme or exposure of the input \cite{xia2020joint,zhang2019deep,zeng2020learning,afifi2021histogan,he2019progressive,wang2019underexposed,gharbi2017deep,he2020condition,wang2021real}. Such a task is different from ours and shares little conception to our current target.  This paper refers to the skin image retouching task aiming to improve skin conditions by removing skin imperfections. Very few techniques exist for face retouching, or specifically, image blemish removal. One direction of the previous works aims to improve the skin image quality by using various filters to smooth out blemishes in images. For instance, local, edge-preserving filters such as Bilateral Filter \cite{tomasi1998bilateral} and Weighted Median Filter \cite{zhang2014100} are applied to interested region{s to smooth the skin}. In contrast, another trend of works (e.g., $L0$ smoothing \cite{xu2011image}, fast global smoothing \cite{min2014fast}) adopts a global smoothing strategy to remove undesired skin details (e.g., blemishes) by energy optimization. Other methods (e.g., Domain Transform \cite{gastal2011domain}, muGIF \cite{guo2017mutually}) smooth images with guided filters to preserve the important structural information. The above-mentioned smoothing methods are designed for general targets. They require manual adjustment of parameters, and often produce under- or over-smoothing effects, failing to achieve a desired skin condition for portrait retouching automatically. 

There exist approaches specifically designed to improve the attractiveness of face photos. Numerous commercial applications achieve skin improvement for portrait retouching tasks, such as Pixlr~\cite{ola2008pixlr}, MeituPic~\cite{meitu2013meitu}, BeautyCam~\cite{beautycam2013meitu}, etc. Despite their function abundance, similar to the traditional approaches, they still require manual efforts to adjust the degree of the applied effect. Lin et al. \cite{lin2019exemplar} proposed an exampler-based method to retouch face images using a small dataset of cosmetic laser therapy. However, their method is limited to blemish removal effects exhibited in laser therapy. 

Two recent CNN-based methods, AutoRetouch \cite{Shafaei2021WACV} and ABPN \cite{lei2022abpn}, share the most similar objectives to ours and aim to remove image imperfections implicitly. AutoRetouch \cite{Shafaei2021WACV} provides a large-scale dataset of manually retouched results based on the FFHQ \cite{karras2019style} dataset, and learns a residual {network} by the paired data, achieving remarkable retouching results. ABPN \cite{lei2022abpn} learns a pyramid of residual maps and blends them back progressively to retouch images in general categories, showing outstanding performances in this task. Despite the impressive results produced by such CNN-based methods, their generalization ability is quite limited, and only produce decent results on the in-domain dataset (Section \ref{exp_generalization}); Moreover, CNN-based methods often remove important details, thus jeopardizing the detail consistency for the non-blemish regions after retouching. Our method provides a stronger solution for out-of-domain data while effectively preserving crucial non-blemish details.

\begin{figure*}
    \centering
    \includegraphics[width=1.0\textwidth]{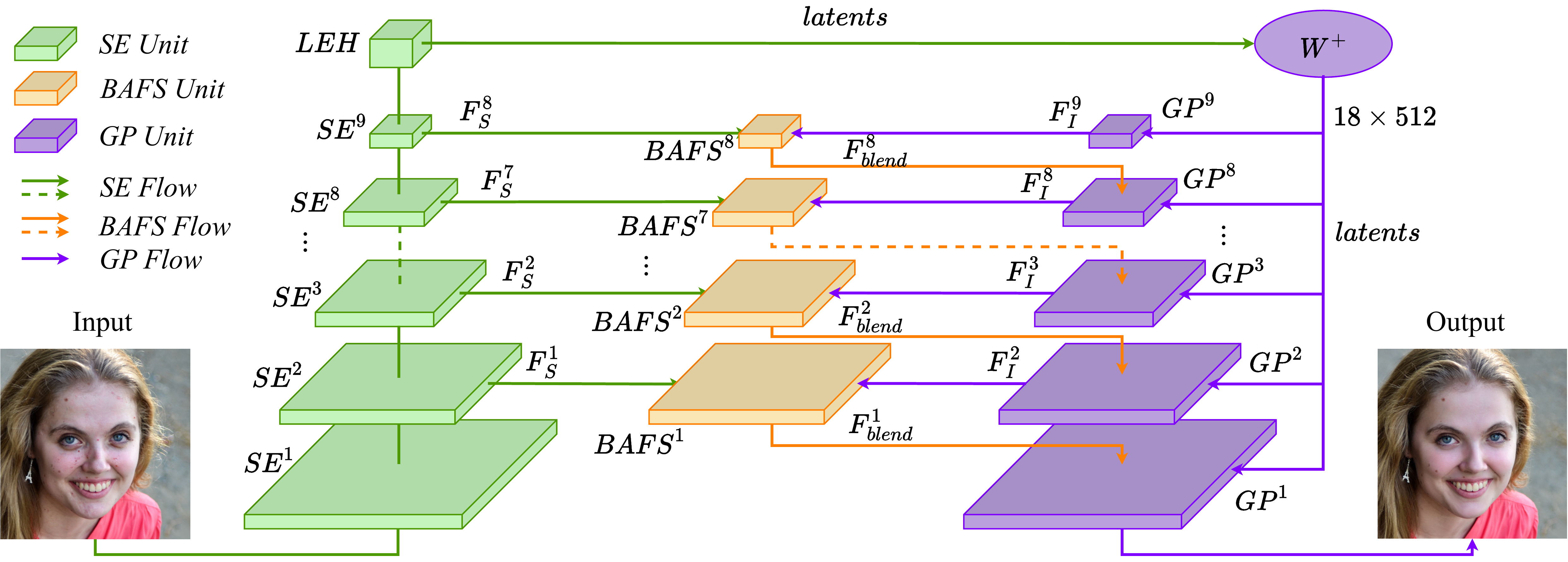}
    \caption{The overall architecture of our \sysName. Given an input image, our semantic extraction module (SE, in green) progressively extracts the semantic features. On the top of the SE units, an LEH unit extracts the features 
    to StyleGAN latent codes in $W^+$ space. Then the latent codes control the generation process of the GAN prior module (GP, in purple). Each GP unit takes a slice of latent code and a blended feature map as input to produce a spatial feature map. The blended feature maps are produced from the {blemish-aware} feature selection (BAFS, in orange) unit at each level by implicitly weighting and blending the feature maps from the input from SE and GP. The fused feature maps are sent back to the next level of the GP units before producing {a} final output image.}
    \label{fig_architecture}
\end{figure*}

\subsection{GAN Prior Incorporation}
Since the advent of GAN \cite{NIPS2014_5ca3e9b1}, various visual problems have been reformulated into the GAN generation process and effectively solved. Among all GAN variants, StyleGAN \cite{karras2019style} generates realistic images with only compact priors as input, motivating many followup works to manipulate the generation with the control of the priors using weights pre-trained on large-scale data. Enormous efforts have been made to obtain the StyleGAN priors, such as optimization-based methods that directly traverse the latent space \cite{abdal2019image2stylegan}, encoder-based methods that map a target input in the latent space \cite{richardson2020encoding,tov2021designing,yang2021gan}, and their hybrid \cite{zhu2020domain}. Our method adopts a similar encoder-based method to obtain StyleGAN priors: we extend the semantic feature processing units with a latent extraction head, converting the spatial feature map to compact priors for the StyleGAN generation process.

One group of previous methods directly manipulate on the priors inverted from the input to achieve image controls\cite{patashnik2021styleclip,harkonen2020ganspace,abdal2021styleflow,gal2022stylegan}. These methods find the GAN priors in a semantically disentangled space and apply semantic controls on the compact priors to further change the content in the output images. Such controls are effective and change the input in semantic aspects, e.g.,  age, gender, hairstyle, pose, lighting, etc. However, controlling the compact priors seems incompetent for tasks requiring strict spatial correspondence or consistency since the image inversion process may incorporate inaccuracy and manipulating the priors directly might further exacerbate such inaccuracy. Thus directly manipulating the latent priors is not feasible for our image retouching task.

Other methods incorporate the GAN priors with intermediate spatial manipulations in the StyleGAN generation process. For example, Image2StyleGAN++ \cite{abdal2020image2styleganpp} blends spatial feature maps from two images to achieve seamless image blending. Wang et al.\cite{wang2021towards} and GLEAN \cite{chan2021glean} incorporate the StyleGAN generation layers as the components in their decoding modules for downstream generation tasks. DrawingInStyles \cite{su2022drawinginstyles} encodes an input to spatial feature maps and bridges it into intermediate layers of the StyleGAN generation process to convert sketches to realistic faces with strict spatial correspondences. The above methods directly operate (e.g., concatenate, replace) on the whole spatial feature maps, without explicitly selecting the features. Our method employs a selective feature blending process, accurately identifying and removing skin blemishes. This precision is challenging to achieve for the mentioned methods that globally operate on feature maps.

{Recently, some methods tend to conduct feature selection to improve the portrait restoration and style transfer problems.} Panini-Net \cite{wang2022panini} fuses the encoded input features with the StyleGAN priors according to channel-wise weights predicted from the input to achieve face restoration. Silimarly, STGAN \cite{liu2019stgan} modifies GRUs \cite{cho2014properties} to selectively preserve the unchanged input  features channel-wisely during generation. VToonify \cite{yang2022vtoonify} predicts a one-channel spatial mask to select the spatial features in fusion, which shares similar idea with one branch of our blending mechanism. According to the statistics in \cite{Shafaei2021WACV}, less than 40\% of the input pixels are changed during retouching, and the alternations occur within a small pixel value range. This implies that for the image retouching task, most pixels in the input should remain intact after editing. To maintain most of the input information unchanged and alter only the blemish regions in the retouching process, we propose a simple yet effective feature selection mechanism (Section \ref{sec_cafb}), utilizing both spatial- and channel-wise attention maps to better balance the composition between the input semantics and GAN priors.

\section{Method}\label{sec_mtd}

\begin{figure*}
    \includegraphics[width=0.98\linewidth]{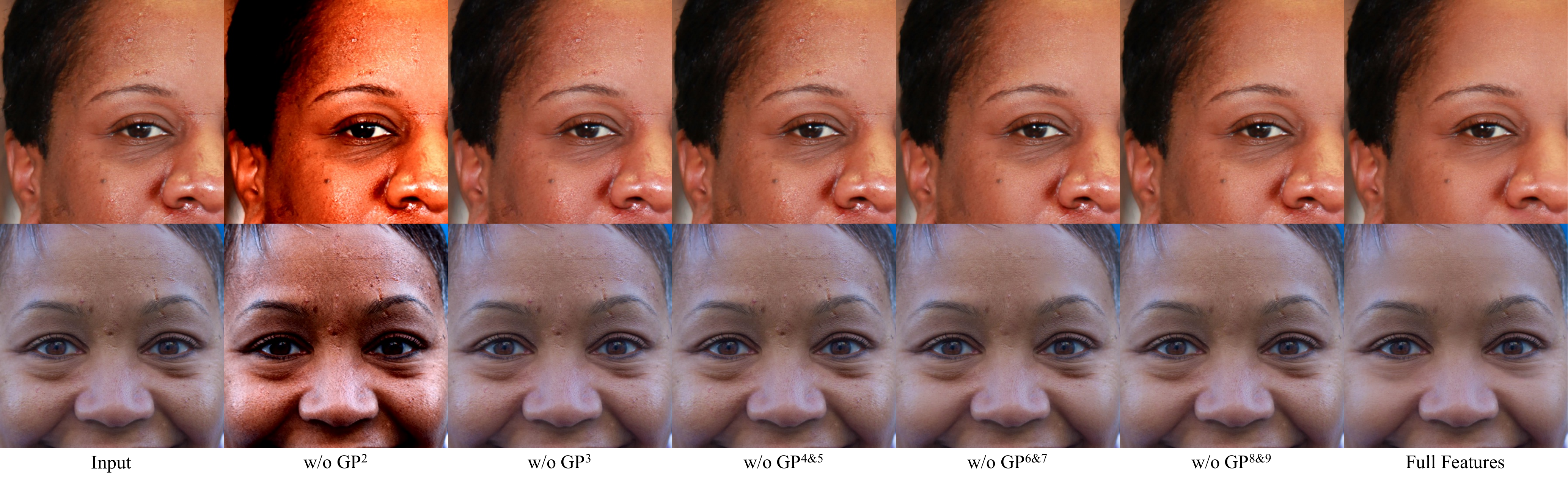}
    \caption{Illustration of the effects of individual levels of GAN prior incorporation. Please zoom-in to find more details for comparison.}
    \label{fig_test_no}
\end{figure*}

\begin{figure}
    \centering
    \includegraphics[width=1.0\linewidth]{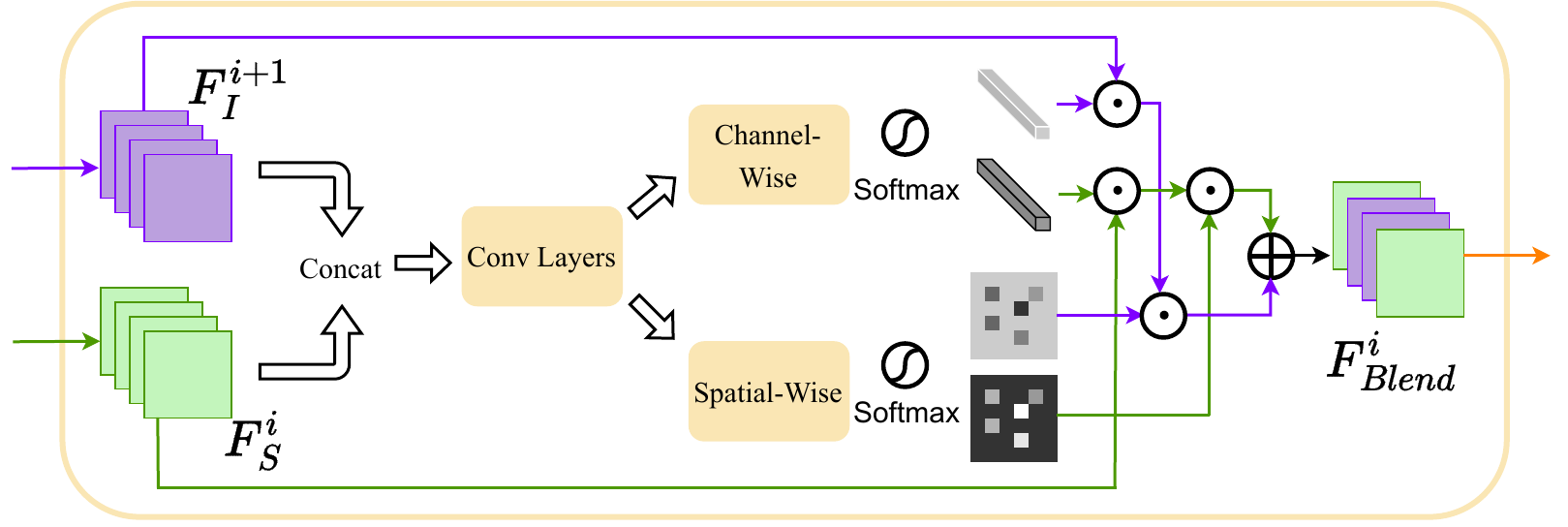}
    \caption{Illustration of a Blemish-Aware Feature Selection (BAFS) unit. The purple and green feature maps represent the intermediate spatial feature maps ${F^{i+1}_{I}}$ from the GP module and spatial semantic feature maps ${F^i_{S}}$ from the SE module, respectively.}
    \label{fig_CAFB}
\end{figure}

Our method consists three main modules: GAN Prior (GP for short, Section \ref{sec_gpp}) module for leveraging pretrained StyleGAN features, Semantic Extraction (SE in abbreviation, Section \ref{sec_sfp}) module to progressively obtain the input features and Blemish-Aware Feature Selection (BAFS, Section \ref{sec_cafb}) module to determine the blemish regions and select the supplementing content in an implicit manner.

\subsection{GAN Prior (GP)}\label{sec_gpp}
Since we want most details of the original input intact, additive to the control of latent codes, we need to inject the spatial features into all the intermediate layers of the StyleGAN to preserve fine details (other than blemishes) in the generated results. Inspired by previous works using StyleGAN layers as the composites of a decoding module \cite{wang2021towards,yang2021gan,chan2021glean}, we bridge the input features to all the intermediate layers of StyleGAN generation layers. We refer to the resulting layers as GP (GAN prior) units in our framework. In the feature infusion process between the input and GAN priors, we propose to implicitly identify and remove skin blemishes by introducing two modalities of masks to adjust the weights for the semantic features. We show the overall framework of our \sysName~in Figure \ref{fig_architecture}, including semantic extraction (SE) module, blemish-aware feature selection (BAFS) module, and GP module. Given an input image $I$, the SE module first extracts its semantic features into a stack of spatial semantic feature maps ${F^i_{S}}, i \in [1, 8]$ and a set of compact priors in the $W+$ space. The GP module takes as input the priors and produces a stack of intermediate spatial feature maps ${F^i_{I}}, i \in [2, 9]$; $GP^1$ unit outputs the final image. The BAFS module dynamically blends the spatial semantic feature maps ${F^i_{S}},$ and the intermediate spatial feature maps ${F^i_{I}}$ at each level. The fused spatial maps are then fed back to the GP module for final result generation. 

\subsection{Semantic Extraction (SE)}\label{sec_sfp}
As we can see in Figure \ref{fig_architecture}, our SE module consists of a stack of $SE^i$ units and an LEH to encode the spatial semantic feature maps ${F^i_{S}}$ and extract the StyleGAN priors in $W+$ space from the input image $I$. 
Mathematically, it is formulated as:
\begin{equation}
    I^i_{I} = 
    \begin{cases}
        SE^i(I), & \text{if} \quad i = 1; \\
        SE^i(I^{i-1}_{I}), & \text{otherwise},
    \end{cases}
    \label{eq_sfp}
\end{equation}
where $\{I^i_{I}\}$ are the intermediate input spatial feature maps. Starting from the 2nd level, each SE unit progressively extracts the semantic information and output{s} $I^i_{I}$ with spatial resolutions reduced by a factor of 2. We append a convolution layer at the end of every SE unit (except the 1st one) to better separate the features between neighboring levels to get $F^i_{S}$:
\begin{equation}
    F^i_{S} = Conv(I^{i+1}_{I}), \quad i \in {[1, 8]},
\end{equation}
where $F^i_{S} \in {\mathbb{R}}^{C_i \times H_i \times W_i}$,
and $H_i = W_i = 2^{10-i}$.
The StyleGAN latent code is extracted at the top of the SE stacks with a Latent Extraction Head (LEH), which contains a convolution and a full-connected layer:
\begin{equation}
    l = LEH(I^{9}_{I}). 
\end{equation}
The latent $l$ is encoded in $W+$ space and of size $\mathbb{R}^{18 \times 512}$.

We adopt a pre-trained StyleGAN2 \cite{Karras2019stylegan2} generator as our GP module to produce the final retouched image. We blend the spatial features of the input and the features $F^i_{I}$ produced by $GP^i$ units before sending the blended features to the subsequent units $GP^{i-1}$. Details of the feature composition will be introduced in Section \ref{sec_cafb}. The progressive generation process is formulated as follows:
\begin{equation}
    F^i_{I} = 
    \begin{cases}
        {GP}^i(const, l^{i}),  & i = 9; \\
        GP^i(F^i_{Blend}, l^{i}), & i \in \bigl[2, 8\bigr],
    \end{cases}
\end{equation}
Each GP unit takes as input the corresponding slices of the encoded latent code $l^i$ as control in the generation process. The $GP^9$ unit takes as input a constant random feature to generate the spatial feature; other $GP^i$ unit takes as input the feature map composited from the counterpart input feature and the feature produced by $GP^{i+1}$ unit in the previous spatial level. $GP^1$ produces the output image of the whole process.

To demonstrate the impact of integrating various levels of GAN priors, we conducted a ablative approach during the generation process. In this process, we intentionally left specific levels $GP^i$ by upsampling and connecting the blended features $F^i_{blend}$ (using BAFS module, elaborate in Section \ref{sec_cafb}) directly to the subsequent StyleGAN blocks $GP^{i-1}$ to produce the output. Figure \ref{fig_test_no} depicts the results of this ablative analysis. The visual depiction highlights that when we remove lower levels of GAN priors, our method yields results with increasingly noticeable artifacts. Higher levels of GAN priors offer finer details and fewer blemishes, indicating their value in progressively guiding our model to detect and correct imperfections across subsequent levels. This contributes to generating refined skin textures in the corresponding regions. Remarkably, when generating images without the inclusion of the $GP^2$ feature, there is a pronounced alteration in color.

\subsection{Blemish-Aware Feature Selection
(BAFS)}\label{sec_cafb}
To enable the implicit blemish identification and removal for the image retouching task, we propose a blemish-aware feature selection module that bridges the SE and GP units in the framework, as illustrated in Figure \ref{fig_architecture}. The composition of $F^i_{Blend}$ with $BAFS^i$ is formulated as:
\begin{equation}
    \begin{split}
        {\mathcal{H}^i} & = {\mathcal{F}}_{Conv}({\mathcal{F}}_{Cat}(F^i_{S},F^{i+1}_{I})), \\
        M^i_C, 1 - M^i_C & = {\mathcal{F}}_{Soft}({\mathcal{F}_{Channel}}({\mathcal{H}^i})), \\
        M^i_S, 1 - M^i_S & = {\mathcal{F}}_{Soft}({\mathcal{F}_{Spatial}}({\mathcal{H}^i})), \\
        F^i_{Blend} & = M^i_S \cdot M^i_C \cdot F^i_{S}, \\
        & + (1 - M^i_S) \cdot (1 - M^i_C) \cdot F^{i+1}_{I},
    \end{split}
    \label{eq_cafb}
\end{equation}
where $\mathcal{F}_{Cat}(\cdot, \cdot)$ represents the concatenation operation in the channel dimension; $\mathcal{F}_{Conv}()$ denotes the common convolution blocks in the unit; $\mathcal{F}_{Channel}$ and $\mathcal{F}_{Spatial}$ represent the channel- and spatial-wise processing, respectively. ${\mathcal{F}}_{Soft}()$ denotes the Softmax operation, and all the multiplication and summation in Equation \ref{eq_cafb} are element-wise operations. See a graphical demonstration of the above process in Figure \ref{fig_CAFB}. With the two separate branches of processing, we finally get two sets of blemish-aware attention masks for the spatial and channel dimensions. $M^i_C$ and $1 - M^i_C$ represent the channel-wise masks at the i-th BAFS unit for GP feature and SE feature, respectively; $M^i_S$ and $1 - M^i_S$ represent the spatial-wise masks at the i-th BAFS unit for GP feature and SE feature, respectively.

\begin{figure}
    \centering
    \includegraphics[width=0.95\linewidth]{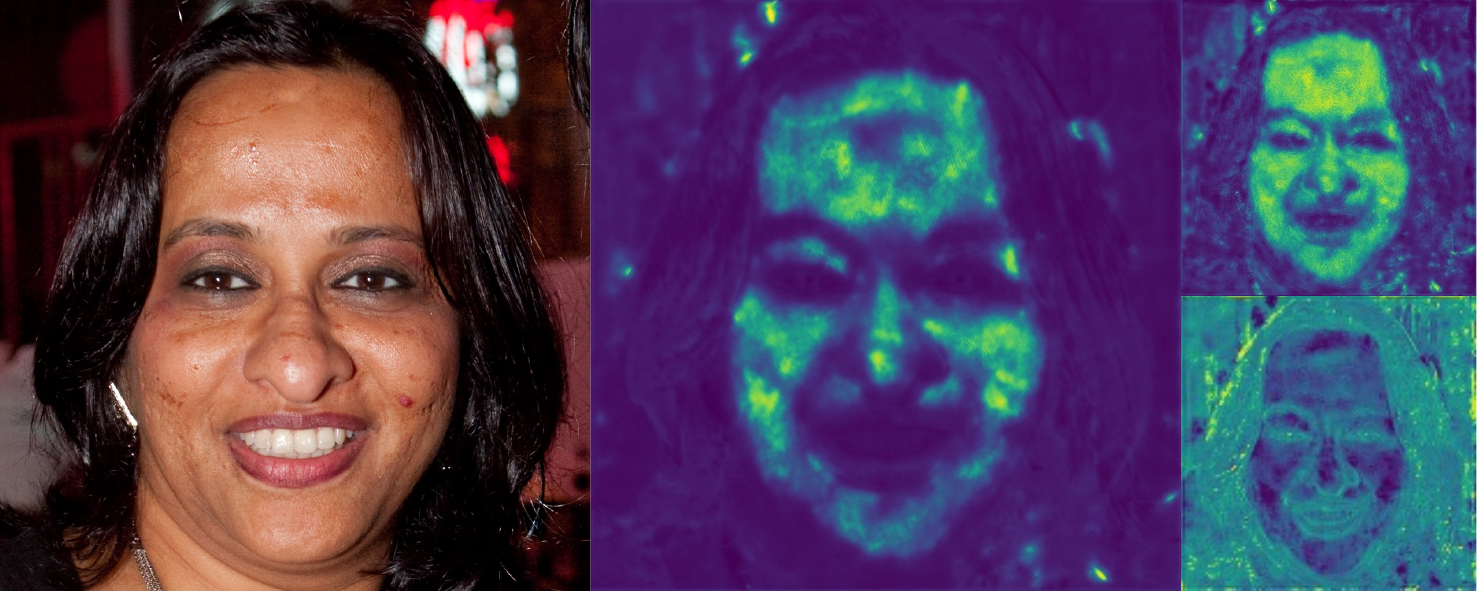}
    \caption{Illustration of {blemish-aware masks ($M^1_S$: middle, $M^2_S$: up-right, and $M^3_S$: down-right)} given an input image.}
    \label{fig_mask}
\end{figure}

We utilize this implicit mask generation mechanism to automatically identify and remove the skin blemishes in the generation process. As there is no ground-truth mask for the blemish area, we do not incorporate pseudo-target as ABPN \cite{lei2022abpn}. Since the blemish regions in face images are usually sparse and disjoint, we cannot apply an accurate threshold to the residual between input and output as a mask for supervision. In the experiment, we can see that our simple selection mechanism is sufficient to identify the blemish regions. We show a representative example in Figure \ref{fig_mask} to demonstrate the blemish-awareness of our spatial masks. The masks with soft boundaries are more suitable for removing the blemishes than the pseudo binary masks. We first load the StyleGAN2 \cite{Karras2019stylegan2} weights pre-trained on the FFHQ dataset \cite{karras2019style} to initialize our GP module and discriminator, and then we finetune the whole pipeline altogether. For the training objectives, we use the $L1$ loss, perceptual loss, and adversarial loss to train our model.

\section{Experiment}
\begin{table}
  \centering
  \small
  \begin{tabular}{ccccc}
    \hline
    Method & Concat & Spatial & Channel & S + C \\
    \hline
    PSNR $\uparrow$ & 37.6300 & 40.9326 & 40.0499 & \textbf{41.0505} \\
    SSIM $\uparrow$ & 0.9725 & \textbf{0.9861} & 0.9837 & 0.9852 \\
    NIQE $\uparrow$ & 4.3112 & 4.2844 & 4.3027 & \textbf{4.3176} \\
    LPIPS $\downarrow$ & 0.0466 & \textbf{0.0413} & 0.0432 & \textbf{0.0413} \\
    \hline
  \end{tabular}
  \caption{Ablation study for different feature blending strategies in the CAFB module. Here ``Concat'', ``Spatial'', and ``Channel'' refer to the results using naive concatenation, spatial-wise blending only, and channel-wise blending only, respectively. ``S + C'' refers to our full method. }
  \label{tab_ablation}
\end{table}

To testify our \sysName's capability on face photo retouching, we conducted extensive evaluations on two types of data: in-domain (experimental) data and out-domain (Internet-collected) data to show its superior performance. We augmented the degrees of blemishes in the original FFHQ dataset to train models capable of processing diverse blemishes as follows. Since the FFHQR \cite{Shafaei2021WACV} contains samples manually retouched from the FFHQ dataset, such paired data gives us a simple way to calculate the blemishes $R^i$ as $R^i = I^i_{F} - I^i_{R}$. Then we multiply a random factor $\lambda \in rand(0,1)$ to blemish $R^i$ before adding it back to the retouched image: $I^i_{aug} = I^i_{R} + \lambda \cdot R^i$ to form a new augmented dataset. We randomly select 10,000 samples for training our model and the other methods for comparison and another 1,000 samples for testing in the following experiments, where the corresponding samples $I^i_{R}$ in the FFHQR dataset are utilized as the ground truth. For evaluations on the generalization abilities of different methods, we manually collected a small set of face images with blemishes from the Internet for qualitative comparison.

\subsection{Ablation Study}\label{exp_ablation}
We first conducted an ablation study on different schemes of feature blending in the BAFS module, including direct concatenation, spatial blending, channel blending, and our spatial + channel blending. For the architecture with no spatial feature blending, the model turns to an encoding-to-style architecture, similar to pSp \cite{richardson2020encoding}. Please refer to the pSp results on quantitative and qualitative evaluations in Section \ref{exp_comp}. We measured several evaluation metrics that are commonly used for image quality assessment, including peak signal-to-noise ratio (PSNR), structural similarity index measure (SSIM)\cite{wang2004image}, natural image quality evaluator (NIQE)\cite{mittal2012making}, and learned perceptual image patch similarity (LPIPS) \cite{zhang2018perceptual}. See Table \ref{tab_ablation} for the detailed values of different blending schemes.

\begin{figure}
    \centering
    \includegraphics[width=1.0\linewidth]{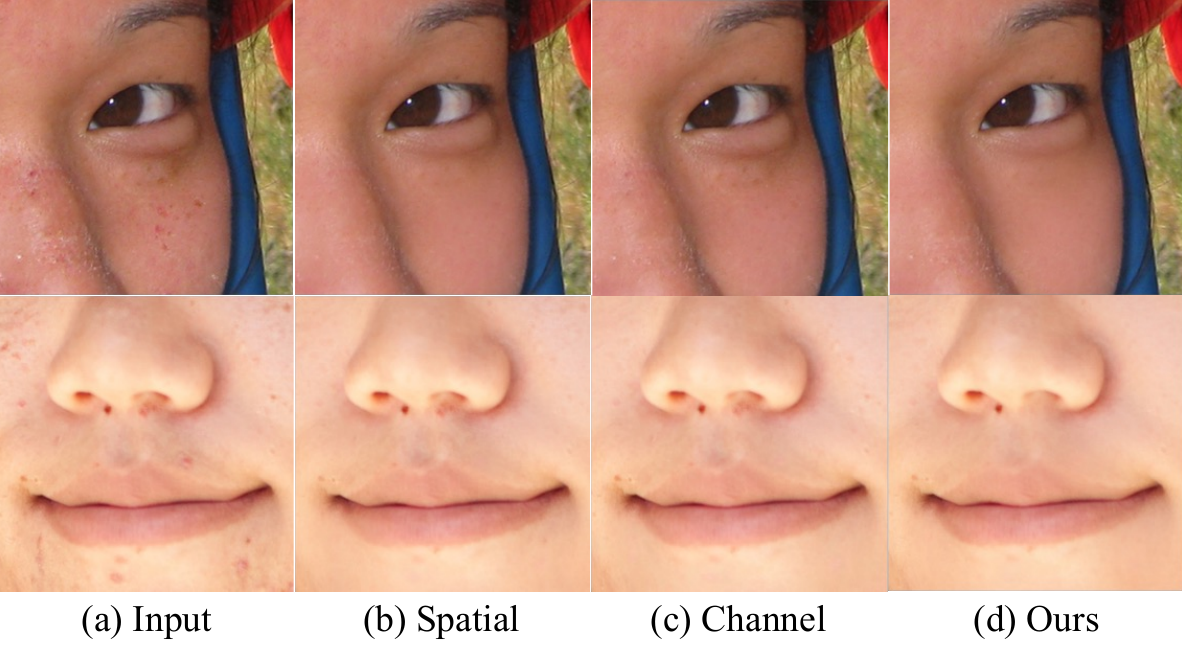}
    
    \caption{Visual comparison in the ablation study. Our choice of incorporating both spatial- and channel-wise blending produces the best retouching results.}
    \label{fig_ablation}
\end{figure}

Except the direct concatenation method, all the metric values reflect quite high performance and show tiny differences among all the indicators even when the retouching results show distinct visual differences (Figure \ref{fig_ablation}). We can see from Figure \ref{fig_ablation} that either spatial blending or channel blending alone produces results inferior to their combination. The spatial blending masks learn to indicate the blemish regions, and the channel blending masks adjust the global weights of the feature blending. The blemishes are handled more precisely by weighting the two mechanisms with these two kinds of masks incorporated in the feature fusion process. We attribute the close quantitative values to the image retouching task itself: Since the retouching operation only affects a small fraction of pixels in the image, the above-mentioned metrics fail to capture the sparse differences, as also confirmed in \cite{Shafaei2021WACV}.

\subsection{Generalization Comparison}\label{exp_generalization}
\begin{figure}
    \centering
    \includegraphics[width=1.1\linewidth]{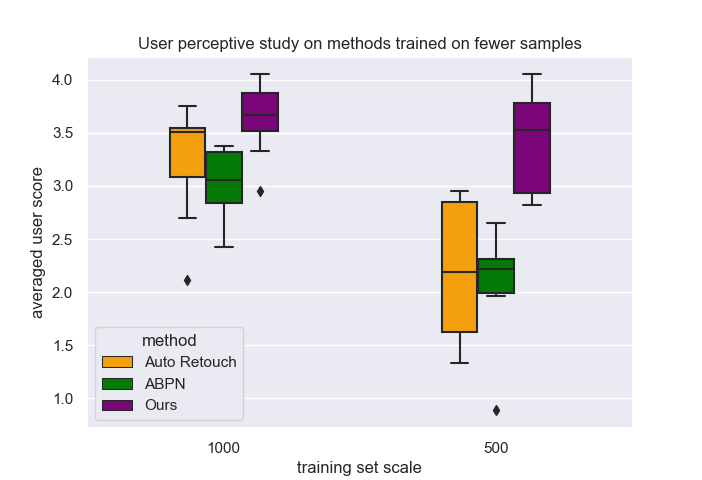}
    \caption{Plot of the statistics for user perceptual study on models with small scale training data.
    The left half and right half illustrate the perceptual result comparisons of the models trained with 1000 and 500 data samples, respectively.}
    \label{fig_user_study_small}
\end{figure}

Given the high cost associated with acquiring manually retouched data for training intricate neural networks, the ability to effectively address the challenges posed by data scarcity becomes imperative. We thus conducted the generalization comparisons to shown our method's ability in the sparse data conditions. To testify the generalization ability of the retouching algorithms, we conducted two experiments: a). training with smaller data scale and b). testing on out-domain data. In our comparative analysis, we opted for two strong baseline methods for comparison: AutoRetouch \cite{Shafaei2021WACV}, a significant competitor for the face retouching task and also the benchmark model from the FFHQR dataset; ABPN \cite{lei2022abpn}, another strong competitor, recently proposed for automatic retouching. Such two methods achieve state-of-the-art performances in the task of learning-based photo retouching. To ensure an fair evaluation, we meticulously reimplemented these two approaches in accordance with their respective descriptions in their papers.

\begin{figure}
    \centering
    \includegraphics[width=\linewidth]{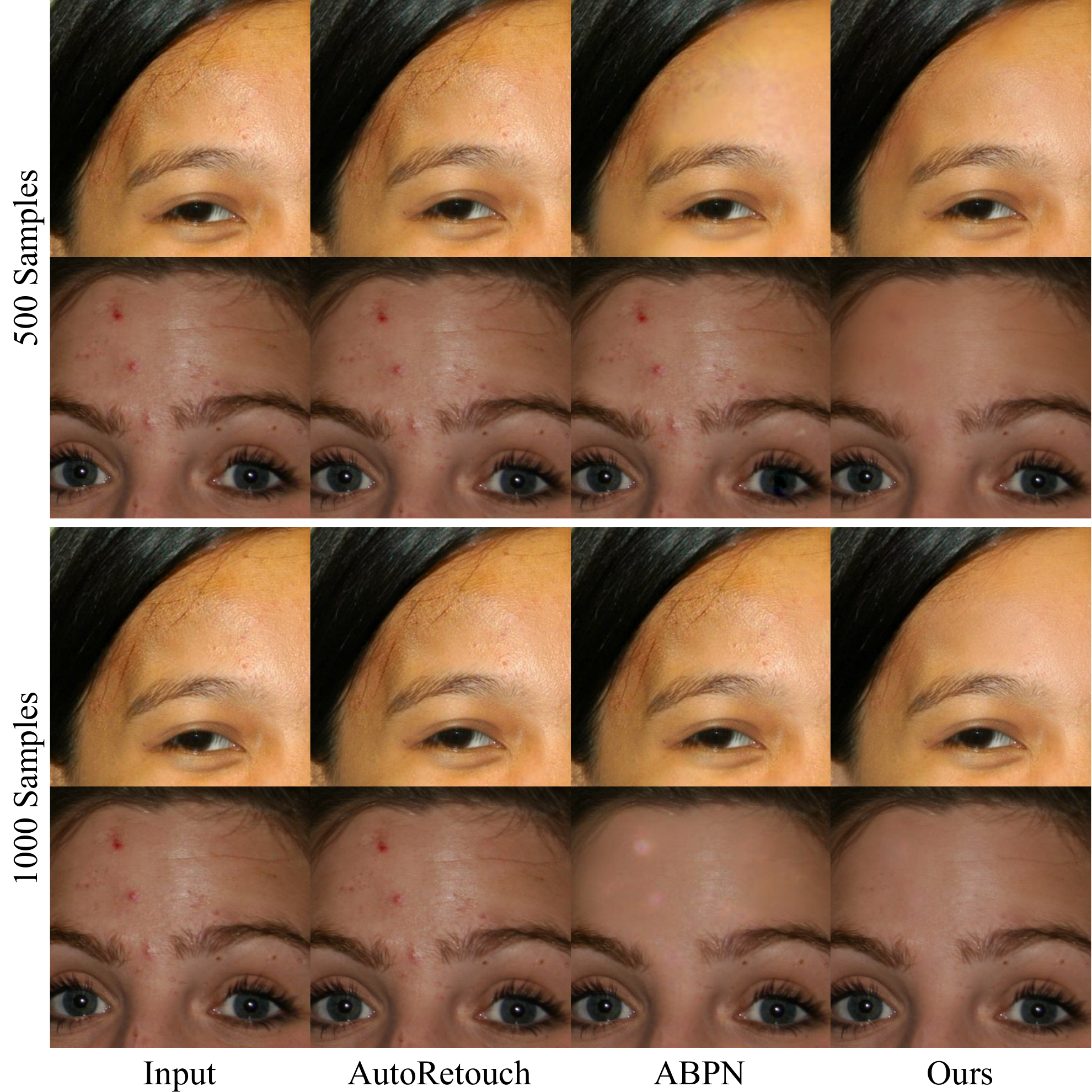}
    \caption{Comparative visual analysis of AutoRetouch, ABPN, and our method trained with small-scale datasets. The top two rows display results from models trained with 500 samples, while the bottom two rows depict results from models trained with 1000 samples.}
    \label{fig_small_data}
\end{figure}

In the first experiment, we reduced the sample numbers in training our method and competing methods by randomly selecting 1,000 and 500 samples from the original training dataset. We tested the models with the original in-domain dataset. We show visual comparisons of this experiment in Figure \ref{fig_small_data}. To further evaluate the visual quality of the retouched results, we conducted a perceptual user study. We compare our method with AutoRetouch and ABPN. We randomly chose 10 inputs from the in-domain dataset and showed study participants with the input and retouched result side-by-side in the evaluation. The results of the three compared methods for a given input were shown in random order {to avoid bias}. The evaluation was conducted via an online questionnaire. 36 participants participated in this study. Each participant was required to evaluate the quality of retouching in a five-point Likert scale (1 = strongly negative to 5 = strongly positive). In total, we got 39 (participants) $\times$ 10 (inputs) = 390 subjective evaluations for each method.

We showed the perceptual score statistics in Figures \ref{fig_user_study_small}. T-tests further show the superior performances of our method: On the 1,000 samples setting: mean: $3.6513$ (ours) vs. $3.2667$ (AutoRetouch), $[t = 3.09, p < 0.01]$; mean: $3.6513$ (ours) vs. $3.0128$ (ABPN), $[t = 4.14, p < 0.01]$. On the 500 samples setting: mean: $3.4000$ (ours)  vs. $2.2051$ (AutoRetouch), $[t = 5.69, p < 0.01]$; mean: $3.4000$ (ours)  vs. $2.1077$ (ABPN), $[t = 5.76, p < 0.01]$. The statistical results conclusively affirm the substantial superiority of our approach when compared to competing methods in terms of retouching quality. Furthermore, user perception results have confirmed that our method consistently delivers reliable retouching results even with a limited training dataset. When the training samples are reduced from 1000 to 500, the average perception score for our method only experiences a modest decrease of $0.2513$, whereas AutoRetouch and ABPN exhibit more substantial score drops of $1.0616$ and $0.9051$, respectively.

In Figure \ref{fig_small_data}, it's evident that AutoRetouch, trained with smaller datasets, generated results almost identical to the inputs, indicating its limited retouching capabilities. As for ABPN, models trained on smaller datasets either closely resemble the inputs or exhibit unnatural textures (as observed in the 1st and 4th rows for ABPN) in the areas with blemishes. Moreover, in our implementation, ABPN models tend to overfit the small dataset quite readily.  Thus, careful selection of the training stopping time becomes imperative; otherwise, the results might encompass extensive areas with unnatural textures, a typical sign of overfitting. In contrast to CNN-based architectures, our method didn’t directly learn the mapping from input to target. Instead, it acted as a bridge between the input and the pre-trained GAN prior space, seeking the feature most akin to the target. This approach enriched the input domain, fostering a wider spectrum of generated variations and preventing it from converging to a single target. As a result, it exhibited superior generalization capabilities in retouching tasks.

\begin{figure}
    \centering
    \includegraphics[width=0.9\linewidth]{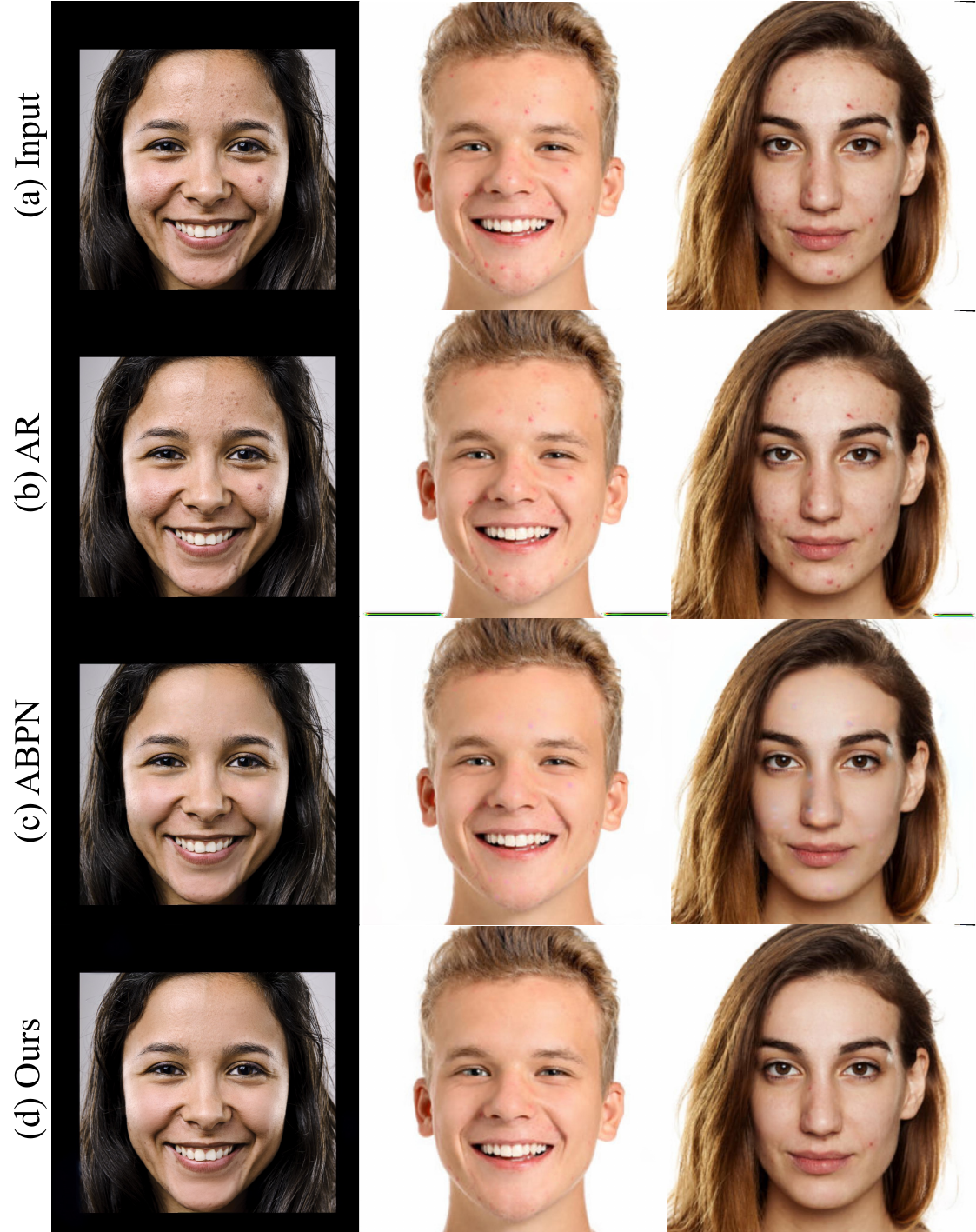}
    \caption{Retouching comparisons on out-domain images between AutoRetouch, ABPN and our method. Please zoom in for more detailed comparisons. }
    \label{fig_comp_out_domain}
\end{figure}

In the second experiment, we applied the models trained on our augmented dataset to out-domain data, i.e., images with skin blemishes collected from the Internet. We collected 35 face photos with blemishes and randomly chose 10 samples processed them with AutoRetouch, ABPN and our method. We evaluated the retouching results with the same user perceptive comparison scheme. See right half of Figure \ref{fig_user_study1} for the resulting statistics. The t-test shows that our method (mean: $3.9308$) significantly outperformed AutoRetouch (mean: $2.1154$; $[t = 17.43, p < 0.01]$) and ABPN (mean: $3.2692$; $[t = 3.26, p < 0.01]$). The visual results in Figure \ref{fig_comp_out_domain} further underscore the superiority of our method. Our approach yielded commendable retouching outcomes, contrasting with AutoRetouch, which closely mirrored the inputs, and ABPN, which introduced unnatural textures. These visual illustrations serve as evidence of our method's superior generalization capabilities compared to AutoRetouch and ABPN when handling out-domain data.

\subsection{Retouch Comparison with Existing Methods}\label{exp_comp}
\begin{figure*}
    \centering
    \includegraphics[width=1.0\linewidth]{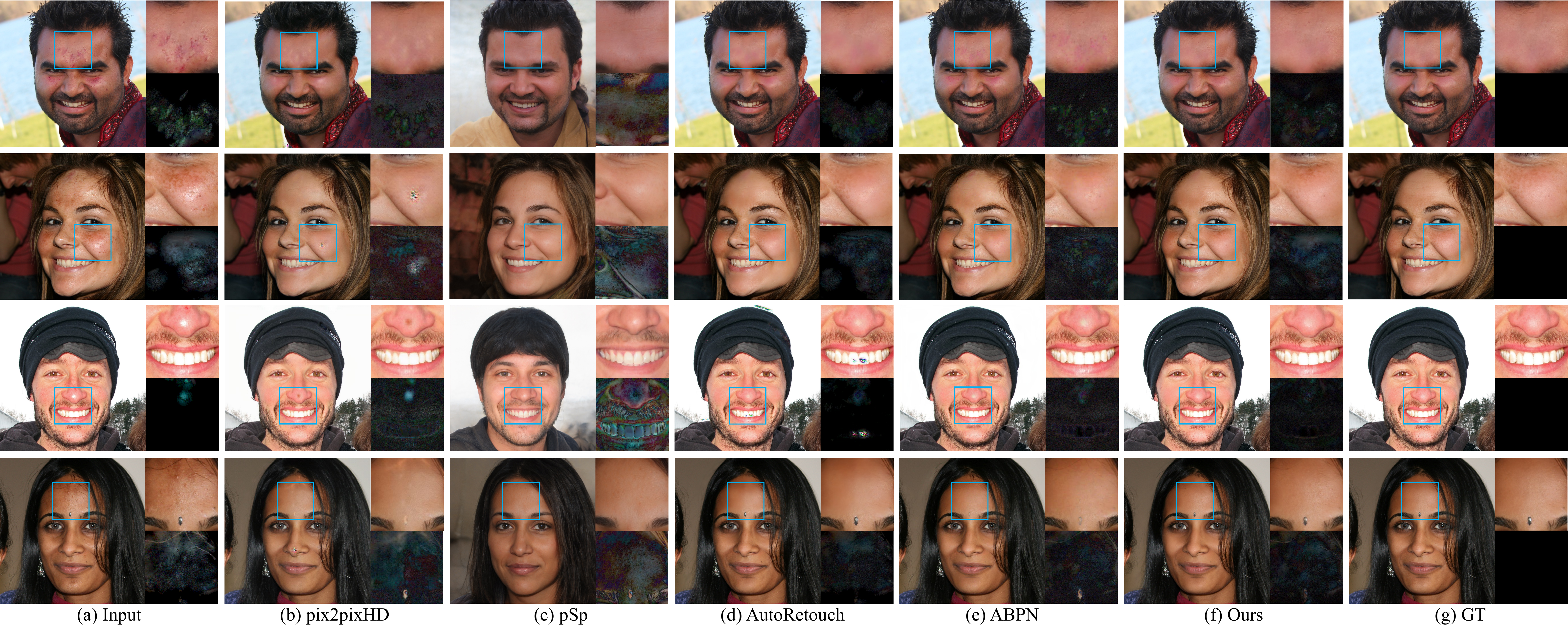}
    \caption{Retouching comparisons by different methods on the FFHQR in-domian dataset. For each sample, we zoom in to show the details of the blue rectangle area on the up-right; and we show the difference map between the result and the ground truth of the zoom-in area on the bottom-right.}
    \label{fig_comparison}
\end{figure*}

\begin{table}
  \centering
  \small
  \begin{tabular}{cccccc}
    \hline
    Method & pixHD & pSp & AR & ABPN & Ours \\
    \hline
    PSNR $\uparrow$ & 30.7053 & 15.7970 & 40.8265 & 39.1939 & 41.0505 \\
    SSIM $\uparrow$ & 0.9104 & 0.6046 & 0.9892 & 0.9466 & 0.9852 \\
    NIQE $\uparrow$ & 4.6528 & 4.4445 & 4.1520 & 4.0508& 4.3176 \\
    LPIPS $\downarrow$ & 0.2150 & 0.5589 & 0.0290 & 0.0736 &0.0413  \\
    \hline
  \end{tabular}
  \caption{Quantitative comparison between our method and alternative methods on the image retouching task. 
  Here ``pixHD'', ``AR'' are abbreviations for ``pix2pixHD'' and ``AutoRetouch'', respectively. }
  \label{tab_comparison}
\end{table}

In addition to AutoRetouch \cite{Shafaei2021WACV} and ABPN\cite{lei2022abpn}, we also compared our method with several state-of-the-art methods for the task of image retouching: pix2pixHD \cite{wang2018pix2pixHD}, a method treating retouching as an image-to-image translation task; pSp \cite{richardson2020encoding}, a comparing method for incorporating StyleGAN with only control in the latent space. We evaluated all the methods using the same training and in-domain evaluation data. We use the same metrics as in Section \ref{exp_ablation} for quantitative evaluation and show the details of the results in Table \ref{tab_comparison}. Figure \ref{fig_comparison} provides a side-by-side comparison of results produced by the compared methods with the difference maps to the ground truths of the highlighted regions for better visual comparison.

\begin{figure}
    \centering
    \includegraphics[width=1.1\linewidth]{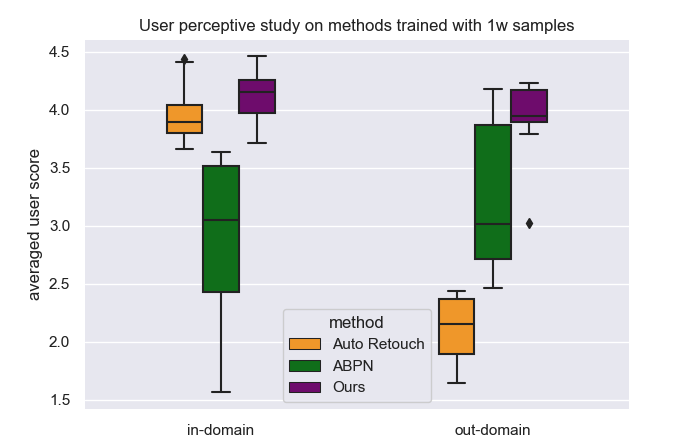}
    \caption{Plot of the statistics for user perceptual study: Left side indicates perceptual scores for three compared methods on in-domain dataset; right side showcases competing perceptual scores on Internet-collected data.}
    \label{fig_user_study1}
\end{figure}

We can see from the visual comparisons that our method provides cleaner retouching and clearer details than all the other methods. Some snow noise-like artifacts exist in some of the pix2pixHD and AutoRetouch results, which are typical artifacts for CNN-based methods (the cheek region of the 4th example in Figure \ref{fig_comparison} (b); and the teeth  area of the 3rd example in Figure \ref{fig_comparison} (d)). As to ABPN, it produced results with homogeneous skin textures and erased some important non-blemish details, such as the decoration in the forehead (only our method and AutoRetouch preserve such detail well.). It also failed to completely remove blemish (1st example) and excessive reflections (3rd example{s} in Figure \ref{fig_comparison} (e)) in the results. Color inconsistency and noisy regions occasionally appear in pix2pixHD results, despite it produced the best NIQE value. pSp exhibited a significant inferior performance in the other three metrics since it cannot preserve the structural information and details compared to the input. 

Since Table \ref{tab_comparison} shows the metric values with no obvious difference, we further incorporate a user perceptive study to evaluate the retouching quality, as in Section \ref{exp_generalization}. On the left side of Figure \ref{fig_user_study1} we show the statistics of the evaluation results. Our method significantly outperformed the two existing methods. Paired t-tests confirmed our method (mean: $4.1282$) produced significantly higher quality retouching results than ABPN (mean: $2.9128$ ; $[t = 5.18, p < 0.01]$). And the mean value of our method is a little bit better than that of AutoRetouch (mean: 3.9743). This results confirmed our method's the superior retouching capability than ABPN and slightly better performance than AutoRetouch with relatively sufficient training data.

\subsection{Retouching Strength Control}
\begin{figure}
    \centering
    \includegraphics[width=1.0\linewidth]{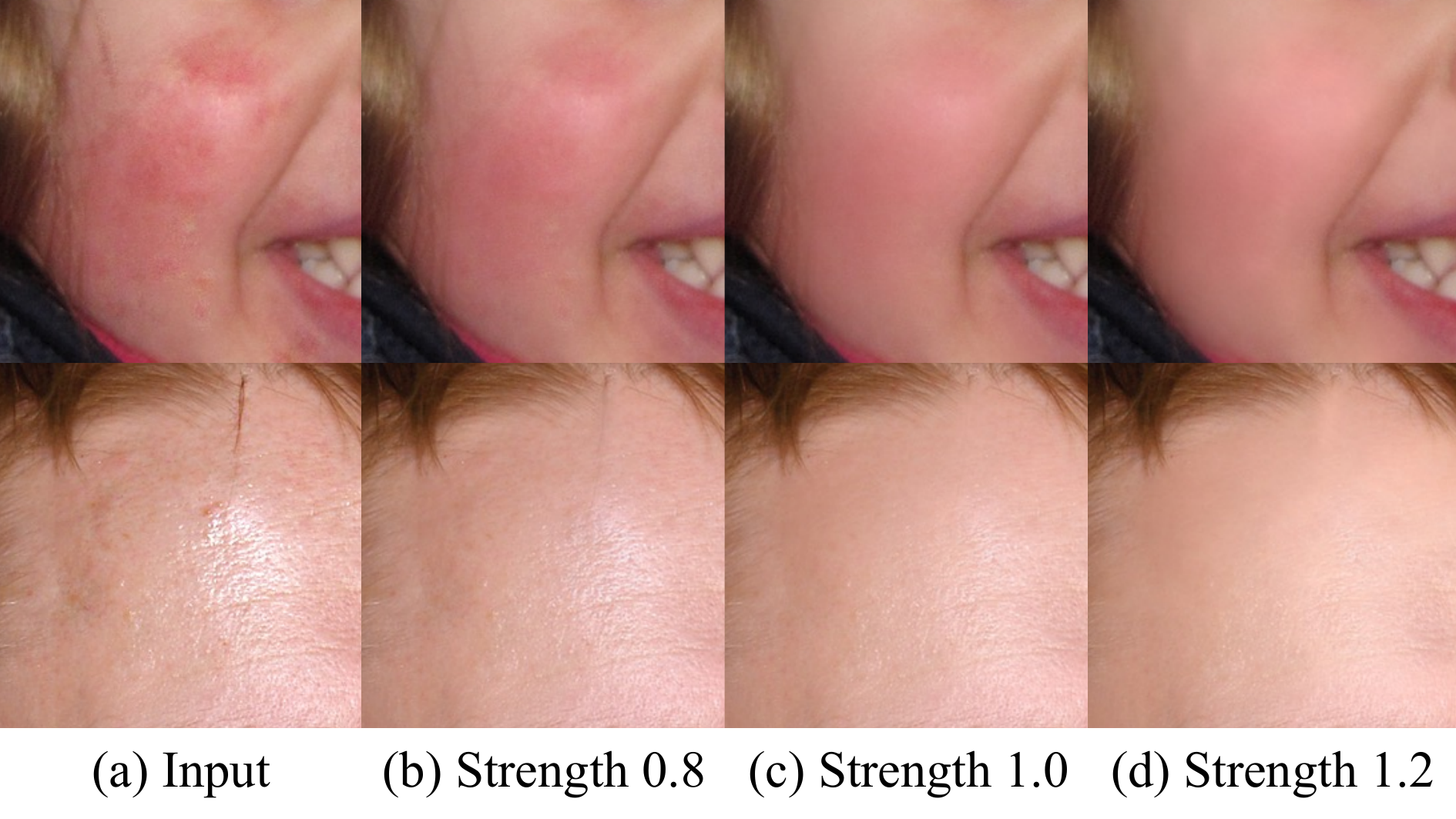}
    \caption{Comparison of retouching strength manipulation. }
    \label{fig_varying}
\end{figure}

Our method supports adjusting the retouching strength in a content-aware manner by changing the channel-wise blending masks in the BAFS module. Different from filter-based methods, which bring global color changes when applying stronger filter weights, our method maintains the non-retouched area intact. Figure \ref{fig_varying} shows two examples of such an effect. In the first example, the reddish face effect diminishes as the strength factor increases; for the second example, the reflections in the forehead region are lightened as the strength factor increases.

StyleGAN is good at synthesizing quality skins, and when adjusting the channel masks, we change the weightings of StyleGAN features and input features in blending. As we keep the spatial masks intact, spatial blending areas remain unchanged as the original retouching; as a result, the weighting changes mainly apply to blemish spots. We adjust the retouching strengths by multiplying a strength factor to the $M^i_C$ mask, $i \in [2, 8]$. By trial-and-error, we found that keeping the first level unchanged and adjusting the other levels, we can achieve the strength-altering results (altering $M^1_C$ would change the global color).

\begin{figure}
    \centering
    \includegraphics[width=0.8\linewidth]{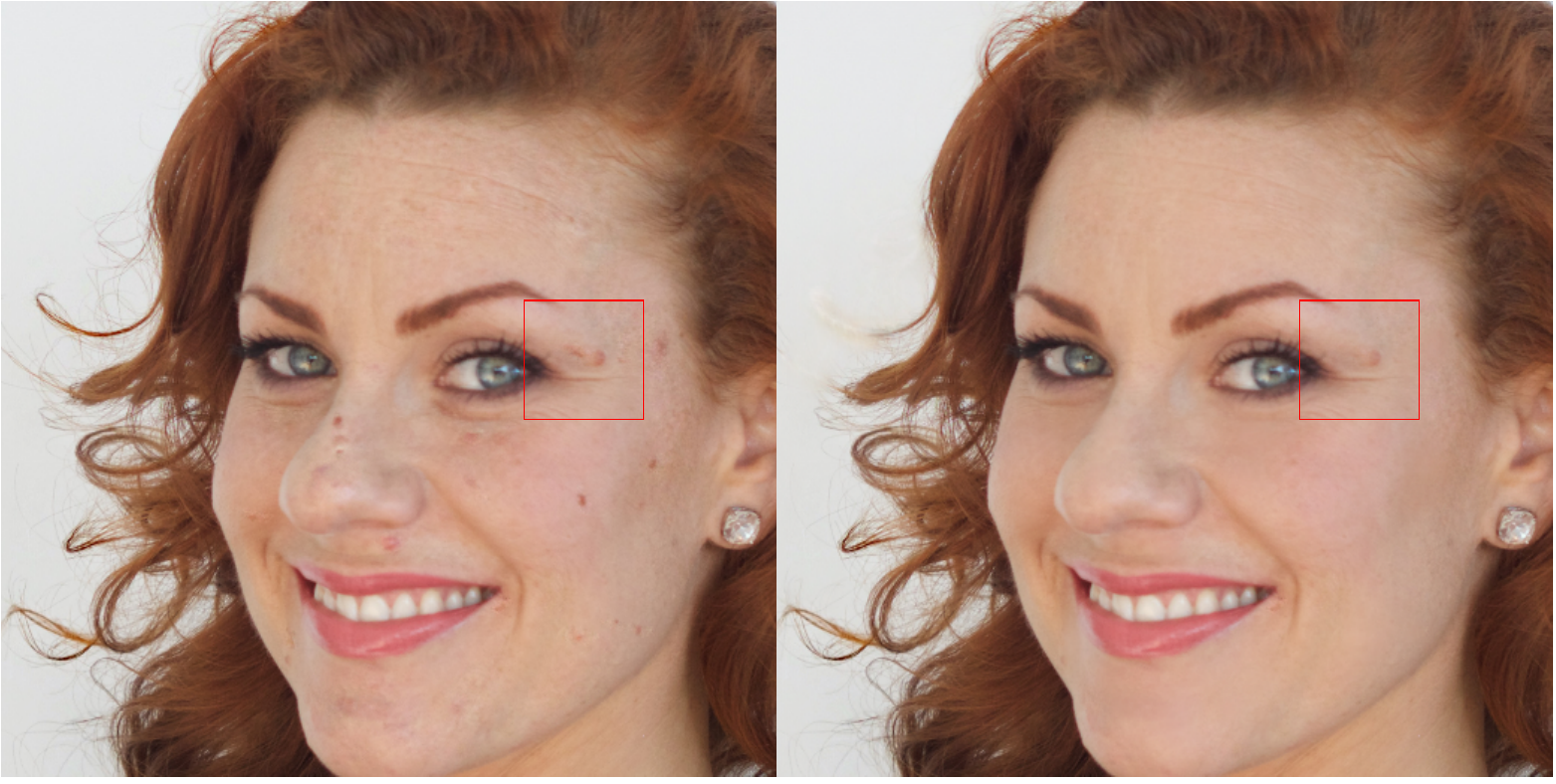}
    \caption{A less successful case of our method failing to remove a blemish due to ambiguity.}
    \label{fig_failure}
\end{figure}

\section{Discussion}
In this paper, we proposed a StyleGAN-based image retouching method, named \sysName, to automatically produce high-quality retouching results for face photos.The utilization of StyleGAN presents a dual-fold advantage: a). It transforms the image-to-image generation problem from a one-to-one mapping process to a nuanced search within the GAN space, allowing for greater flexibility in image generation; b). The pretrained StyleGAN demonstrates proficiency in generating specific skin features essential in the retouching process. Due to the incorporation of StyleGAN priors, our method shows superior generalization capability compared to conventional CNN-based solutions. Our blemish-aware feature selection mechanism enables our method to identify and remove blemishes implicitly from the supervision of paired data. The cascaded infusion of input features effectively preserves the fine details in the retouching process.

Despite the good results our method produced, our method fails to remove some blemishes completely, as shown in Figure \ref{fig_failure}. It is sometimes ambiguous to tell whether an area is a blemish or not, {making our method keep the spot intact}. Another limitation is that our current implementation is limited to the face photo retouching task since the StyleGAN model is trained on a face dataset. A general-targeted image retouching method may still need further efforts.

With cascaded infusion of input features to the StyleGAN intermediates, our method produces results with strong spatial correspondence. Similar tasks could also benefit from our design, such as inpainting, relighting, etc. In our quantitative evaluations, we found that the popular image quality assessment metrics fail to faithfully reflect the image retouching quality. {We thus depended on user perceptive study to assess the retouching quality in our experiments.} A reliable measurement to evaluate the sparse image editing quality would greatly improve current image retouching performances. Efforts toward this direction would also be promising.

{\small
\bibliographystyle{ieee_fullname}
\bibliography{ref_file}
}

\end{document}